\documentclass{article}

\usepackage[final]{corl_2026}   

\usepackage{amsmath,amssymb,amsfonts}
\usepackage{graphicx}
\usepackage{booktabs}
\usepackage{multirow}
\usepackage{xcolor}
\usepackage{microtype}
\usepackage{subcaption}
\usepackage{tabularx}
\usepackage{array}
\usepackage{placeins}

\makeatletter
\renewcommand{\@noticestring}{CoRL 2026 Workshop: Do Robots Need World Models?}
\makeatother

\newcommand{\tabfm}{TabFM}
\newcommand{\tabpfn}{TabPFN}
\newcommand{\Enc}{E}
\newcommand{\Dec}{D}
\newcommand{\R}{\mathbb{R}}
\newcommand{\E}[2]{$#1\mathrm{e}{#2}$}
\newcommand{\EB}[2]{$\mathbf{#1}\mathbf{e}\mathbf{#2}$}
\newcommand{\new}[1]{#1} 

\title{An Evaluation of Tabular Foundation Models as Training-Free World Models}

\author{
  Xian Yeow Lee, Ryoichi Takase, Ahmed Farahat\\
  Industrial AI Lab\\
  Hitachi America, Ltd.\\
  \texttt{\{xian.lee, ryoichi.takase, ahmed.farahat\}@hal.hitachi.com}
}

\begin{document}
\maketitle

\begin{abstract}
World models are widely considered to be a potential key component of model-based reinforcement learning and robot planning, but typically require task-specific training on substantial interaction data. We investigate whether a pretrained tabular foundation model (\tabfm) can serve as a training-free dynamics model for world modeling by framing next-state prediction as in-context regression over transition pairs, and where such a model helps and where it breaks. Across synthetic dynamical systems and robot locomotion benchmarks, \tabfm~is more accurate than LSTM and MLP models trained from scratch on the same limited transitions in almost all settings\new{, and among regressors fitted to the same data it is the most accurate on locomotion, while a Gaussian process is more accurate on smooth low-dimensional systems}. However, models trained on the full dataset remain substantially better on most systems. When rolled out autoregressively over multiple steps, \tabfm's accuracy depends strongly on how much context it is given. With a short context, its rollouts quickly degrade to the level of a trivial baseline, whereas with a long context they improve substantially and, on one locomotion task, exceed a trained LSTM. Inference is also about three orders of magnitude slower than an LSTM. Overall, in their current state, we find that training-free priors are useful \new{as predictors} when data are scarce and enough context is available, but they are not yet a substitute for trained world models in planning.
\end{abstract}


\section{Introduction}
\label{sec:intro}

World models are widely studied as a way to support model-based reinforcement learning, robot planning, and physical AI systems~\citep{worldmodels,wang2026world}. Approaches such as DreamerV3~\citep{dreamerv3}, TD-MPC2~\citep{tdmpc2}, and RSSM~\citep{rssm} learn dynamics from environment interactions via gradient descent, often requiring substantial task-specific data for each new deployment. We ask a different question: can a pretrained foundation model provide a useful prior for next-state prediction without task-specific dynamics training, and if so, what are its limits?

Tabular foundation models are a natural candidate for this role. Models such as \tabpfn~\citep{tabpfnv2} are pretrained on datasets generated by Structural Causal Models (SCMs) and perform in-context regression on a provided set of input-output pairs. Since SCMs describe relationships of the form $x_j = f_j(\text{pa}(x_j), \varepsilon_j)$, and physical state transitions $x_{t+1} = f(x_t, a_t, \varepsilon_t)$ can be expressed in a structurally similar form, we hypothesize that the SCM prior transfers to next-state prediction. Unlike conventional world models, a \tabfm~can then exploit additional transition data at inference time simply by extending its context window, without any gradient updates.

However, a world model that is useful for a robot needs more than accurate one-step prediction; it has to support multi-step rollouts for planning and remain competitive with models trained on large datasets. To offer a perspective on whether robots need world models, in particular how far a general-purpose prior can go without task-specific training, we ask not only whether a \tabfm~can predict the next state, but also \emph{where such a training-free prior helps and where it breaks}. Our contributions are:
\begin{itemize}
  \item We formulate dynamics prediction as tabular in-context regression and evaluate an off-the-shelf \tabfm~as a training-free dynamics predictor.
  \item \textbf{Where it helps:} given the same limited data, \tabfm~is more accurate than neural networks trained from scratch and, \new{for locomotion tasks, than classical regressors such as a Gaussian process}. With a long context, its multi-step rollouts improve substantially.
  \item \textbf{Where it breaks:} \new{a Gaussian process is more accurate on smooth low-dimensional systems,} trained models remain substantially more accurate when data are available, short-context rollouts degrade to the level of a trivial baseline, and inference is orders of magnitude slower.
\end{itemize}

\section{Related Work}
\label{sec:related}

\paragraph{In-context dynamics prediction:} Closest to our work, \citet{benechehab2025zeroshot} use a pretrained LLM in-context to predict the dynamics of continuous control tasks, including D4RL HalfCheetah and Hopper, and project states onto decorrelated principal components, predicting each component independently. The One-Shot World Model (OSWM)~\citep{ferreira2024oneshot} trains a prior-data fitted network (PFN)~\citep{pfn} on a synthetic dynamics prior and uses it in-context as a world model on simple control tasks. Other work pretrains dynamics models on synthetic systems~\citep{forgione2023system,ziegler2024foundation} and shows that in-context prediction error decreases with context length~\citep{liu2024llms,zhang2025zeroshot}. \tabpfn~\citep{tabpfnv2,pfn} itself has been extended to time-series forecasting~\citep{tabpfnts,jayawardhana2026zeroshot} and used as a gradient-free Q-function~\citep{icrrl,tabql}. In contrast, we use a general-purpose tabular model pretrained on SCM data, rather than a language model or a dynamics-specific prior, off-the-shelf for action-conditioned next-state prediction\new{, testing whether a prior never trained on dynamics transfers to them}.

\paragraph{World models and the one-step objective:} Model-based RL methods~\citep{chua2018deep,janner2019trust,dreamerv3,tdmpc2} learn dynamics with gradient descent on environment-specific data. It is well known that one-step accuracy is an imperfect proxy for what a planner needs. Errors compound over rollouts~\citep{talvitie2014model,venkatraman2015improving}, and model likelihood can correlate poorly with control performance~\citep{lambert2020objective}. We extend this direction of research to tabular foundation models, and find that for these models, the performance depends strongly on context length.

\section{Method}
\label{sec:method}

\paragraph{Framing dynamics as tabular in-context regression:}
Let $x_t \in \R^d$ be the state at time $t$. For action-conditioned systems, we augment the state with the action $a_t \in \R^k$: $\tilde{x}_t = [x_t \,|\, a_t] \in \R^{d+k}$. The model is then asked to predict the next state $x_{t+1} \in \R^d$. Given a history of $H$ consecutive transitions, we construct an in-context dataset
\begin{equation}
  \mathcal{C} = \{(\tilde{x}_{\tau}, \Delta x_{\tau})\}_{\tau=t-H}^{t-1}, \qquad \Delta x_\tau = x_{\tau+1} - x_\tau .
\end{equation}
Each pair becomes one tabular row with input features $\tilde{x}_\tau$ and regression target $\Delta x_\tau$. Given the query $\tilde{x}_t$, the \tabfm~predicts $\widehat{\Delta x}_t$, and the next-state estimate is $\hat{x}_{t+1} = x_t + \widehat{\Delta x}_t$. Since \tabpfn~is a scalar regressor, we issue $d$ sequential calls per timestep, one per state dimension. All input dimensions remain visible as features in each call, but cross-dimension correlations in the output are not modeled. 

\paragraph{Direct and latent pipelines:} We consider two ways of using \tabfm~as a world model. In the \emph{direct} pipeline, $\tilde{x}_t$ is used directly as a tabular row. Since the direct pipeline does not scale to high-dimensional observations such as images, we also consider a \emph{latent} pipeline that applies the \tabfm~to latent transitions $\{(z_\tau, \Delta z_\tau)\}$ with $z_t = \Enc(x_t)$, where the encoder-decoder is trained per environment with a reconstruction and temporal smoothness objective (Appendix~\ref{app:details}). Because the encoder requires task-specific training, the latent pipeline is not training-free and is excluded from the matched-budget comparison experiments below.

\section{Experimental Setup}
\label{sec:setup}

\paragraph{Datasets:}
We evaluate on three low-dimensional systems, namely two autonomous systems (Lorenz and a linear dynamical system, LDS) and an action-conditioned damped pendulum, and on the D4RL Hopper and HalfCheetah medium datasets~\citep{d4rl}. For the low-dimensional systems, we collect a single 5{,}000-step trajectory and split it 80:20 into training and evaluation. For D4RL, we hold out 10 episodes for evaluation. The training split is used only by the trained baselines (Appendix Table~\ref{tab:datasets}).

\paragraph{Baselines:}
We compare \tabfm~against baselines that see the same context window, and against models trained on the full training split. \new{Five} baselines are training-free: \emph{persistence}, which predicts no change ($\hat{x}_{t+1} = x_t$), \emph{ridge} regression\new{, a Gaussian process (GP), kernel ridge regression and $k$-nearest neighbors ($k$-NN), all} fitted to the same normalized rows. Two are trained from scratch on the window: LSTM-few-shot and MLP-few-shot are initialized fresh at every evaluation step and fitted on the same $H$ transitions with up to 200 gradient steps. Finally, as data-rich references, we train a two-layer LSTM and an MLP that takes the flattened window of past transitions as input on the full training split. All errors are computed based on the state vector only. Actions are treated as inputs, and rollouts use the recorded actions. \new{During rollouts, the context is a sliding window that takes in each predicted state, and all context lengths share the same start points.} We report one-step MSE for context lengths from $H{=}5$ to $500$, and the error after 20 autoregressive steps at $H \in \{20, 100, 500\}$. \tabpfn~v2~\citep{tabpfnv2} is used off-the-shelf with default hyperparameters (Appendix~\ref{app:details}).

\section{Results}
\label{sec:results}

\subsection{Where It Helps: In-Context Inference under Matched Data Budgets}
\label{sec:fewshot}

\begin{figure}[t]
  \centering
  \includegraphics[width=0.93\linewidth]{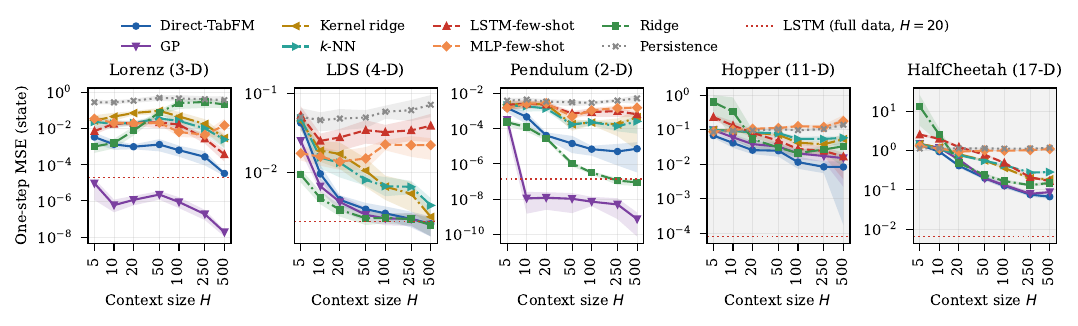}
  \vspace{-0.7em}
  \caption{\textbf{In-context inference vs.\ other uses of the same $H$ transitions.} All methods receive the same context window; dotted: LSTM trained on full data. Bands: 95\% bootstrap intervals.}
  \label{fig:fewshot}
  \vspace{-1em}
\end{figure}

\begin{table}[t]
\centering
\caption{One-step state MSE\new{. Context-only methods use their full context ($H{=}500$; 250 on Hopper); the} LSTM and MLP\new{, trained on the full data, use the 20-step window they were trained with}. Bold: within 5\% of the lowest.}
\label{tab:besth}
\scriptsize
\setlength{\tabcolsep}{2.2pt}
\begin{tabular}{lccccccccc}
\toprule
Dataset & \new{Persistence} & Ridge & \new{GP} & \new{Kernel ridge} & \new{$k$-NN} & Direct-\tabfm & Latent-\tabfm & LSTM & MLP \\
\midrule
Lorenz & \new{\E{3.7}{-1}} & \new{\E{2.1}{-1}} & \new{\EB{1.9}{-8}} & \new{\E{3.0}{-3}} & \new{\E{2.4}{-3}} & \new{\E{3.3}{-5}} & \new{\E{1.0}{-2}} & \new{\E{2.0}{-5}} & \new{\E{1.4}{-4}} \\
LDS & \new{\E{7.2}{-2}} & \new{\EB{2.1}{-3}} & \new{\EB{2.2}{-3}} & \new{\E{2.7}{-3}} & \new{\E{3.8}{-3}} & \new{\EB{2.2}{-3}} & \new{\EB{2.2}{-3}} & \new{\E{2.4}{-3}} & \new{\E{3.5}{-3}} \\
Pendulum & \new{\E{5.2}{-3}} & \new{\E{9.0}{-8}} & \new{\EB{7.4}{-10}} & \new{\E{4.5}{-4}} & \new{\E{2.6}{-4}} & \new{\E{7.3}{-6}} & \new{\E{9.5}{-5}} & \new{\E{1.4}{-7}} & \new{\E{1.2}{-5}} \\
\midrule
Hopper & \new{\E{1.0}{-1}} & \new{\E{2.7}{-2}} & \new{\E{1.7}{-2}} & \new{\E{3.7}{-2}} & \new{\E{5.3}{-2}} & \new{\E{8.3}{-3}} & \new{\E{1.7}{-2}} & \new{\EB{8.0}{-5}} & \new{\E{2.7}{-4}} \\
HalfCheetah & \new{\E{1.2}{0}} & \new{\E{1.5}{-1}} & \new{\E{8.8}{-2}} & \new{\E{1.9}{-1}} & \new{\E{2.8}{-1}} & \new{\E{6.7}{-2}} & \new{\E{1.3}{-1}} & \new{\EB{6.6}{-3}} & \new{\E{1.6}{-2}} \\
\bottomrule
\end{tabular}
\vspace{-0.5em}
\end{table}

We first isolate the effect of the pretraining prior by giving all methods the same transitions (Figure~\ref{fig:fewshot}). For $H \geq 10$, \tabfm~is more accurate than both few-shot LSTM and MLP on every system, by up to two orders of magnitude. From $H{=}20$ on, \tabfm's advantage over the few-shot baselines is consistent across evaluation queries on every system (Appendix Table~\ref{tab:boot}). Ridge regression on the same window is a stronger baseline: it is more accurate than \tabfm~on LDS, which is linear, and on Pendulum, which is nearly linear at the small angles in our data. On the nonlinear systems the order reverses: \tabfm~is more accurate than ridge on Lorenz beyond the shortest context and on both locomotion tasks at every $H$, and its error keeps decreasing with more context. \new{Kernel ridge and $k$-NN are less accurate than \tabfm~on every system for $H \geq 10$, but the GP is far more accurate on the smooth Lorenz and Pendulum systems, while \tabfm~matches or beats it on locomotion.}

\subsection{Comparison with Fully Trained Models}
\label{sec:fulldata}

Table~\ref{tab:besth} \new{compares all methods at fixed context lengths.} With at most 500 transitions, \tabfm~comes close to the fully trained LSTM on Lorenz and matches it on LDS. On Pendulum, ridge and the trained LSTM are both more accurate, consistent with its near-linear dynamics. \new{On Lorenz and Pendulum, the GP is more accurate than even the trained LSTM.} On D4RL locomotion, where the baselines are trained on about 800K transitions, the trained models remain one to two orders of magnitude more accurate. Latent-\tabfm~is less accurate than Direct-\tabfm~except on LDS.

\subsection{Multi-Step Rollouts Depend on Context}
\label{sec:rollout}

\begin{table}[t]
\centering
\caption{20-step rollout state MSE \new{from the same start points}. Bold: within 5\% of the lowest.}
\label{tab:rollout}
\scriptsize
\setlength{\tabcolsep}{1.9pt}
\begin{tabular}{lcc|ccc|ccc|ccc}
\toprule
& \multicolumn{2}{c|}{\new{any $H$}} & \multicolumn{3}{c|}{\new{$H{=}20$}} & \multicolumn{3}{c|}{\new{$H{=}100$}} & \multicolumn{3}{c}{\new{$H{=}500$}} \\
Dataset & \new{Pers.} & \new{LSTM} & \new{Ridge} & \new{GP} & \new{\tabfm} & \new{Ridge} & \new{GP} & \new{\tabfm} & \new{Ridge} & \new{GP} & \new{\tabfm} \\
\midrule
Lorenz & \new{\E{9.3}{1}} & \new{\E{7.8}{-3}} & \new{\E{2.0}{2}} & \new{\E{1.5}{1}} & \new{\E{1.2}{2}} & \new{\E{4.9}{2}} & \new{\E{5.3}{-1}} & \new{\E{4.7}{1}} & \new{\E{8.4}{1}} & \new{\EB{5.9}{-6}} & \new{\E{1.4}{-1}} \\
LDS & \new{\E{2.5}{-2}} & \new{\E{1.9}{-2}} & \new{\E{1.8}{-2}} & \new{\E{3.4}{-2}} & \new{\E{5.1}{-2}} & \new{\E{1.8}{-2}} & \new{\E{2.1}{-2}} & \new{\E{3.3}{-2}} & \new{\EB{1.6}{-2}} & \new{\EB{1.6}{-2}} & \new{\E{2.2}{-2}} \\
Pendulum & \new{\E{4.6}{-1}} & \new{\E{2.4}{-5}} & \new{\E{1.3}{-1}} & \new{\E{2.2}{-5}} & \new{\E{6.2}{-1}} & \new{\E{1.3}{-4}} & \new{\E{8.6}{-7}} & \new{\E{8.4}{-4}} & \new{\E{7.7}{-6}} & \new{\EB{2.0}{-7}} & \new{\E{2.8}{-4}} \\
Hopper & \new{\E{7.6}{0}} & \new{\EB{9.0}{-3}} & \new{\E{9.7}{1}} & \new{\E{2.2}{1}} & \new{\E{2.6}{1}} & \new{\E{6.0}{1}} & \new{\E{1.0}{1}} & \new{\E{3.5}{0}} & \new{\E{3.5}{0}} & \new{\E{1.3}{1}} & \new{\E{2.0}{0}} \\
HalfCheetah & \new{\E{1.9}{0}} & \new{\E{7.5}{-1}} & \new{\E{1.7}{2}} & \new{\E{5.9}{0}} & \new{\E{4.5}{0}} & \new{\E{4.1}{-1}} & \new{\E{3.1}{-1}} & \new{\E{2.9}{-1}} & \new{\E{3.4}{-1}} & \new{\E{3.2}{-1}} & \new{\EB{2.5}{-1}} \\
\bottomrule
\end{tabular}
\vspace{-1em}
\end{table}

Table~\ref{tab:rollout} shows that rollout quality depends strongly on context length. With a short context ($H{=}20$), \tabfm's 20-step rollouts are no more accurate than persistence on any system, even though its one-step predictions are more accurate. \new{From the same start points, extending the context to $H{=}500$ reduces \tabfm's rollout error by more than an order of magnitude on every nonlinear system, bringing it below both persistence and ridge on Lorenz and both locomotion tasks.} The trained LSTM, which needs only a 20-step window, remains one to two orders of magnitude more accurate on Lorenz and Hopper, is statistically indistinguishable on LDS and Pendulum, and is less accurate on HalfCheetah. \new{The GP's rollouts are the most accurate on Lorenz and Pendulum, but trail \tabfm's on Hopper at long context.} A likely explanation is that \new{with a short window, the context soon consists mostly of the model's own predictions, so its errors feed back into the in-context data, whereas a long window remains dominated by real transitions.} Because \tabfm~reprocesses the context for every prediction and state dimension, a single prediction on Hopper takes about 3\,s, \new{about 1000$\times$} slower than the trained LSTM\new{, though a GP is slower still at $H{=}500$} (Appendix Table~\ref{tab:latency}).

\section{Discussion and Conclusion}
\label{sec:discussion}
Our results suggest three conclusions. First, a prior that has never seen robot data extracts more from a few transitions than learning from scratch \new{and, on locomotion, than a Gaussian process, which is more accurate on smooth low-dimensional systems}, but fully trained models remain substantially more accurate when enough data are available. Second, one-step accuracy at a fixed context is not sufficient evidence that a model is useful for planning. With a short context, \tabfm~beats persistence in one step but not after 20 steps; with a long context, its rollouts improve but still trail a short-window trained LSTM on Lorenz and Hopper. Evaluations should therefore report rollouts across context lengths alongside simple baselines. Third, given its inference cost, a training-free predictor may currently suit offline uses such as data augmentation or policy evaluation better than the inner loop of a planner\new{, as well as settings where retraining is impractical}. In this work, we evaluate explicit predictors only. Future work includes closed-loop and pixel-based evaluation, comparisons with dynamics-specific priors, experiments across multiple seeds, and reducing inference cost. 


\bibliography{corl_references}

\clearpage
\appendix
\section*{Supplementary Material}
\addcontentsline{toc}{section}{Supplementary Material}

\section{Experimental Details}
\label{app:details}

\paragraph{Datasets.}
Table~\ref{tab:datasets} summarizes the datasets and their properties. For each low-dimensional system, we collect a single trajectory of 5{,}000 steps (seed 0) and hold out the last 20\% for evaluation; training data is used exclusively to train baselines, while \tabfm~is evaluated on the eval split without any training. Pendulum is a custom damped pendulum integrated with the Euler method, with state $[\theta, \omega]$ and Gaussian torque inputs with standard deviation 1; on this trajectory the pole stays at small angles ($|\theta| < 0.5$\,rad), where the dynamics are close to linear. For D4RL, we use the medium-quality offline datasets~\citep{d4rl}, split them into episodes at terminals and timeouts before any window is built, train baselines on the full training split, and evaluate on 10 held-out episodes for tractable \tabpfn~inference. We also ran Gymnasium CartPole-v1, but excluded it from this version: under a random policy its episodes are short, and our current windowing rule, which keeps each context within one episode, leaves almost no valid windows at $H \geq 50$.

\paragraph{One-step evaluation.}
All errors are computed on the state dimensions only; actions are model inputs and are never scored. One-step MSE is averaged over 50 query indices per evaluation trajectory, drawn with replacement with a fixed seed, so each low-dimensional system has 50 queries and each D4RL dataset up to 500. All methods share the same query indices. On Hopper at $H{=}500$, only 3 of the 10 evaluation episodes are long enough to host the context window, so that column uses fewer queries ($^\dagger$ in Table~\ref{tab:e2_d4rl})\new{, and Table~\ref{tab:besth} uses $H{=}250$ for Hopper instead}.

\paragraph{Rollout evaluation.}
Rollouts start from 50 points in the evaluation data (5 per episode for D4RL), feed the predicted state back while keeping the recorded actions, and report the state MSE at the 20th step. \new{The context is a sliding window of length $H$: after each step, the oldest row is dropped and the query is appended with its state replaced by the prediction, so from the third step on the context contains predicted states. All methods and context lengths use the same start points, namely those with at least 500 prior transitions.} The LSTM and MLP use their own 20-step input window from the same start points. On Hopper, only one evaluation episode is long enough \new{for these start points (36 starts)}, so no confidence interval is reported and these Hopper results are indicative only.

\paragraph{Confidence intervals.}
Since we use a single data and training seed, the intervals reflect variability over evaluation queries, not over seeds. To compare two methods, we resample the evaluation queries with replacement, compute the ratio of their mean errors on the resampled set, and repeat this 4{,}000 times; the middle 95\% of these ratios gives the interval. The comparison is paired, meaning both methods are always scored on the same resampled queries. If the whole interval lies below 1, \tabfm~is more accurate regardless of which subset of queries is drawn. Because neighbouring queries are correlated, we resample whole episodes for D4RL and 100-step blocks of the single trajectory for the low-dimensional systems. The bands in the figures are computed in the same way.

\begin{table}[h]
\centering
\caption{\textbf{Dataset summary.}
\tabfm~uses no task-specific training on any dataset. LDS denotes Linear Dynamical Systems.}
\label{tab:datasets}
\small
\begin{tabular}{llrrrr}
\toprule
Environment & Source & Train steps & State dim & Action dim & Type \\
\midrule
Lorenz        & Synthetic & 4{,}000       & 3   & --- & Autonomous \\
LDS           & Synthetic & 4{,}000       & 4   & --- & Autonomous \\
Pendulum      & Synthetic & 4{,}000       & 2   & 1   & Controlled \\
\midrule
Hopper        & D4RL      & 797{,}732     & 11  & 3   & Controlled \\
HalfCheetah   & D4RL      & 800{,}000     & 17  & 6   & Controlled \\
\bottomrule
\end{tabular}
\end{table}

\paragraph{Baselines.}
We compare against \new{nine} baselines:
\begin{itemize}
  \item \textbf{Persistence}: predicts $\hat{x}_{t+1} = x_t$.
  \item \textbf{Ridge}: a linear ridge regression ($\alpha{=}1$) fitted in closed form to the same normalized context rows as \tabfm, predicting the state residual.
  \item \new{\textbf{GP}: Gaussian process regression with kernel $c \cdot \mathrm{RBF}(\ell) + \mathrm{White}(\sigma^2)$, a single length scale and normalized targets; $c$, $\ell$ and $\sigma^2$ are fitted by maximizing the marginal likelihood on the context rows, with two optimizer restarts.}
  \item \new{\textbf{Kernel ridge}: kernel ridge regression with an RBF kernel, $\alpha{=}1$ and $\gamma{=}1/d$, where $d$ is the number of input features.}
  \item \new{\textbf{$k$-NN}: distance-weighted $k$-nearest-neighbor regression with $k{=}\min(5, H)$. Like ridge, the GP, kernel ridge and $k$-NN use the same normalized context rows as \tabfm, fit one regressor per state dimension, and have no hyperparameters tuned on evaluation data.}
  \item \textbf{LSTM}: a two-layer LSTM autoregressive predictor trained on the full training split with supervision on the next-state residual. It is trained with a fixed context length of 20.
  \item \textbf{MLP}: a context-window MLP that flattens the context window and maps to $x_{t+1}$, also trained on the full training split. Each MLP entry in Appendix~\ref{app:results} is a separately trained model at the corresponding context length.
  \item \textbf{LSTM-few-shot}: a fresh-initialized LSTM adapted on the exact context window used by \tabfm~at each evaluation step, with up to 200 gradient steps and early stopping. No pretraining is used; this provides a data-budget-matched comparison against \tabfm. Models are reinitialized at every step.
  \item \textbf{MLP-few-shot}: identical protocol to LSTM-few-shot but using a fresh MLP. Its first layer takes the flattened context window concatenated with the query, i.e.\ $(H{+}1)$ times the row dimension, so its input grows with context length while the number of adaptation samples stays small. This baseline is therefore structurally disadvantaged at large $H$; the architecture of LSTM-few-shot does not depend on $H$.
\end{itemize}
All learned baselines use residual prediction. The fully trained LSTM and MLP and the few-shot baselines are all trained with the loss on the state only. Latent-\tabfm~is excluded from the few-shot comparison since its encoder is trained on the full dataset, introducing additional supervised learning not present in the few-shot baselines.

\paragraph{Latent pipeline.}
The latent pipeline compresses observations through a learned representation before applying the \tabfm:
\begin{align}
  z_t &= \Enc(x_t) \in \R^{d_z}, \qquad
  \hat{x}_{t+1} = \Dec(\hat{z}_{t+1}),
\end{align}
where $\hat{z}_{t+1}$ is the \tabfm's prediction in latent space. The encoder-decoder is trained with the combined objective
\begin{equation}
  \mathcal{L} = \|x - \Dec(\Enc(x))\|^2 + \lambda \|\Enc(x_{t+1}) - \Enc(x_t)\|^2 .
\end{equation}
The first term minimizes reconstruction error, while the second encourages temporal smoothness in the latent space by keeping consecutive latent states close. The encoder is trained per environment and independently of the \tabfm, without gradients flowing through the tabular model at any point; dynamics prediction is then handled entirely by the pretrained tabular prior. Note that for the low-dimensional systems, $d_z{=}16$ is larger than the state dimension, so the encoder acts as a learned re-parameterization rather than a compression.

\paragraph{Implementation.}
\tabpfn~v2~\citep{tabpfnv2} is used off-the-shelf with default hyperparameters and no fine-tuning on any of the evaluation datasets. Before each call, the context window is z-score normalized per feature and the query is normalized with the same statistics; the output is then de-normalized. For proprioceptive observations, the autoencoder consists of a two-layer MLP encoder and decoder with hidden width 64 and latent dimension $d_z = 16$, trained for 50 epochs with Adam at a learning rate of $10^{-3}$ and temporal smoothness coefficient $\lambda = 0.1$. All experiments are conducted on a single NVIDIA RTX 4090 GPU\new{, except the GP, kernel ridge and $k$-NN baselines, which use scikit-learn on CPU}.

\FloatBarrier
\section{Full Numerical Results}
\label{app:results}

\begin{figure}[h]
  \centering
  \includegraphics[width=\linewidth]{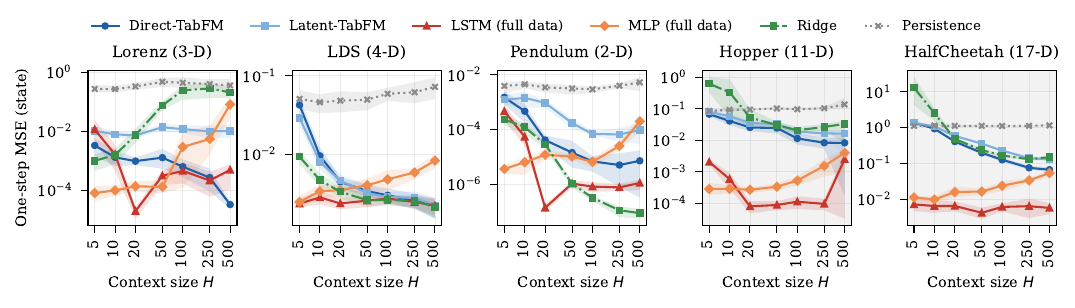}
  \caption{\textbf{One-step state MSE vs.\ context length (log-log), full-data setting.} Direct-\tabfm~improves with context length, while the LSTM and MLP are best near their training configuration. Bands: 95\% bootstrap intervals over evaluation queries.}
  \label{fig:scaling}
\end{figure}

Figure~\ref{fig:scaling} shows the one-step state MSE of all methods across context lengths, and Tables~\ref{tab:e2_synthetic} and~\ref{tab:e2_d4rl} report the corresponding numbers. LSTM is trained with a fixed context length of 20 and receives longer sequences out-of-distribution at evaluation, while each MLP entry represents a separately trained model at the corresponding context length. Table~\ref{tab:boot} reports paired bootstrap ratios for the main comparisons, Table~\ref{tab:rollfull} reports rollouts of all methods across context lengths \new{from the same start points}, and Table~\ref{tab:latency} reports inference times.

\begin{table}[h]
\centering
\caption{\textbf{One-step state MSE on the low-dimensional systems.} Bold = lowest per dataset and context length.}
\label{tab:e2_synthetic}
\scriptsize
\setlength{\tabcolsep}{3pt}
\begin{tabular}{llccccccc}
\toprule
Dataset & Model & 5 & 10 & 20 & 50 & 100 & 250 & 500 \\
\midrule
\multirow{11}{*}{Lorenz}
  & Persistence & 0.28 & 0.28 & 0.34 & 0.49 & 0.45 & 0.41 & 0.37 \\
  & Ridge & \E{1.0}{-3} & \E{1.6}{-3} & \E{7.5}{-3} & 0.078 & 0.25 & 0.29 & 0.21 \\
  & \new{GP} & \new{\EB{9.4}{-6}} & \new{\EB{5.8}{-7}} & \new{\EB{1.1}{-6}} & \new{\EB{2.2}{-6}} & \new{\EB{8.7}{-7}} & \new{\EB{1.9}{-7}} & \new{\EB{1.9}{-8}} \\
  & \new{Kernel ridge} & \new{0.028} & \new{0.046} & \new{0.072} & \new{0.094} & \new{0.046} & \new{0.018} & \new{\E{3.0}{-3}} \\
  & \new{$k$-NN} & \new{0.023} & \new{0.018} & \new{0.020} & \new{0.036} & \new{0.026} & \new{\E{1.0}{-2}} & \new{\E{2.4}{-3}} \\
  & Direct-\tabfm & \E{3.4}{-3} & \E{1.3}{-3} & \E{9.8}{-4} & \E{1.3}{-3} & \E{6.4}{-4} & \E{2.7}{-4} & \E{3.3}{-5} \\
  & Latent-\tabfm & 0.010 & \E{8.0}{-3} & \E{7.4}{-3} & 0.014 & 0.012 & 0.010 & 0.010 \\
  & LSTM & 0.012 & \E{1.7}{-3} & \E{2.0}{-5} & \E{3.1}{-4} & \E{4.6}{-4} & \E{2.2}{-4} & \E{5.0}{-4} \\
  & MLP & \E{8.1}{-5} & \E{9.9}{-5} & \E{1.4}{-4} & \E{1.3}{-4} & \E{3.0}{-3} & \E{5.6}{-3} & 0.083 \\
  & LSTM-few-shot & \E{7.2}{-3} & 0.018 & 0.020 & 0.022 & 0.016 & \E{2.8}{-3} & \E{3.7}{-4} \\
  & MLP-few-shot & 0.034 & 0.021 & 0.019 & 0.021 & \E{6.4}{-3} & \E{4.5}{-3} & 0.015 \\
\midrule
\multirow{11}{*}{LDS}
  & Persistence & 0.051 & 0.046 & 0.048 & 0.050 & 0.059 & 0.062 & 0.072 \\
  & Ridge & \E{9.3}{-3} & \E{4.7}{-3} & \E{3.3}{-3} & \E{2.6}{-3} & \E{2.6}{-3} & \E{2.5}{-3} & \EB{2.1}{-3} \\
  & \new{GP} & \new{0.025} & \new{\E{6.6}{-3}} & \new{\E{4.2}{-3}} & \new{\E{2.9}{-3}} & \new{\EB{2.6}{-3}} & \new{\E{2.5}{-3}} & \new{\E{2.2}{-3}} \\
  & \new{Kernel ridge} & \new{0.045} & \new{0.019} & \new{0.017} & \new{0.010} & \new{\E{6.6}{-3}} & \new{\E{5.3}{-3}} & \new{\E{2.7}{-3}} \\
  & \new{$k$-NN} & \new{0.048} & \new{0.015} & \new{0.013} & \new{\E{8.0}{-3}} & \new{\E{6.6}{-3}} & \new{\E{6.5}{-3}} & \new{\E{3.8}{-3}} \\
  & Direct-\tabfm & 0.042 & \E{9.5}{-3} & \E{4.5}{-3} & \E{3.4}{-3} & \E{3.0}{-3} & \E{2.5}{-3} & \E{2.2}{-3} \\
  & Latent-\tabfm & 0.029 & \E{8.0}{-3} & \E{4.6}{-3} & \E{3.2}{-3} & \E{2.9}{-3} & \E{2.8}{-3} & \E{2.2}{-3} \\
  & LSTM & \EB{2.3}{-3} & \EB{2.8}{-3} & \EB{2.4}{-3} & \EB{2.6}{-3} & \E{2.7}{-3} & \EB{2.5}{-3} & \E{2.2}{-3} \\
  & MLP & \E{2.5}{-3} & \E{3.4}{-3} & \E{3.5}{-3} & \E{4.0}{-3} & \E{4.8}{-3} & \E{5.9}{-3} & \E{8.3}{-3} \\
  & LSTM-few-shot & 0.053 & 0.025 & 0.028 & 0.034 & 0.032 & 0.034 & 0.039 \\
  & MLP-few-shot & 0.017 & 0.016 & 0.014 & 0.015 & 0.023 & 0.022 & 0.022 \\
\midrule
\multirow{11}{*}{Pendulum}
  & Persistence & \E{3.9}{-3} & \E{4.5}{-3} & \E{3.5}{-3} & \E{3.1}{-3} & \E{2.9}{-3} & \E{3.9}{-3} & \E{5.2}{-3} \\
  & Ridge & \E{2.3}{-4} & \E{1.2}{-4} & \E{2.9}{-5} & \E{1.1}{-6} & \E{3.2}{-7} & \E{1.1}{-7} & \E{9.0}{-8} \\
  & \new{GP} & \new{\E{3.2}{-4}} & \new{\EB{1.1}{-8}} & \new{\EB{1.2}{-8}} & \new{\EB{1.1}{-8}} & \new{\EB{7.2}{-9}} & \new{\EB{5.1}{-9}} & \new{\EB{7.4}{-10}} \\
  & \new{Kernel ridge} & \new{\E{2.3}{-3}} & \new{\E{2.6}{-3}} & \new{\E{2.4}{-3}} & \new{\E{1.8}{-4}} & \new{\E{2.2}{-4}} & \new{\E{1.8}{-4}} & \new{\E{4.5}{-4}} \\
  & \new{$k$-NN} & \new{\E{2.1}{-3}} & \new{\E{1.9}{-3}} & \new{\E{1.4}{-3}} & \new{\E{1.7}{-4}} & \new{\E{2.3}{-4}} & \new{\E{1.4}{-4}} & \new{\E{2.6}{-4}} \\
  & Direct-\tabfm & \E{1.5}{-3} & \E{4.7}{-4} & \E{4.1}{-5} & \E{1.4}{-5} & \E{6.9}{-6} & \E{5.1}{-6} & \E{7.3}{-6} \\
  & Latent-\tabfm & \E{1.2}{-3} & \E{1.4}{-3} & \E{9.4}{-4} & \E{1.7}{-4} & \E{6.9}{-5} & \E{6.7}{-5} & \E{9.5}{-5} \\
  & LSTM & \E{4.6}{-4} & \E{5.4}{-5} & \E{1.4}{-7} & \E{1.1}{-6} & \E{8.3}{-7} & \E{7.8}{-7} & \E{1.1}{-6} \\
  & MLP & \EB{3.7}{-6} & \E{6.4}{-6} & \E{1.2}{-5} & \E{1.0}{-5} & \E{6.7}{-6} & \E{2.5}{-5} & \E{2.0}{-4} \\
  & LSTM-few-shot & \E{2.3}{-3} & \E{2.7}{-3} & \E{2.3}{-3} & \E{7.4}{-4} & \E{8.8}{-4} & \E{9.9}{-4} & \E{6.5}{-4} \\
  & MLP-few-shot & \E{1.5}{-3} & \E{2.6}{-3} & \E{2.8}{-3} & \E{4.9}{-4} & \E{1.1}{-3} & \E{1.5}{-3} & \E{1.6}{-3} \\
\bottomrule
\end{tabular}
\end{table}

\begin{table}[h]
\centering
\caption{\textbf{One-step state MSE on D4RL locomotion.} Bold = lowest per dataset and context length. $^\dagger$ = based on 76 queries from the 3 longest episodes (Appendix~\ref{app:details}).}
\label{tab:e2_d4rl}
\scriptsize
\setlength{\tabcolsep}{3pt}
\begin{tabular}{llccccccc}
\toprule
Dataset & Model & 5 & 10 & 20 & 50 & 100 & 250 & 500 \\
\midrule
\multirow{11}{*}{Hopper}
  & Persistence & 0.085 & 0.092 & 0.094 & 0.10 & 0.096 & 0.10 & 0.14$^\dagger$ \\
  & Ridge & 0.62 & 0.33 & 0.053 & 0.030 & 0.021 & 0.027 & 0.033$^\dagger$ \\
  & \new{GP} & \new{0.096} & \new{0.060} & \new{0.038} & \new{0.032} & \new{0.020} & \new{0.017} & \new{0.015$^\dagger$} \\
  & \new{Kernel ridge} & \new{0.098} & \new{0.093} & \new{0.081} & \new{0.071} & \new{0.042} & \new{0.037} & \new{0.053$^\dagger$} \\
  & \new{$k$-NN} & \new{0.096} & \new{0.093} & \new{0.082} & \new{0.078} & \new{0.054} & \new{0.053} & \new{0.058$^\dagger$} \\
  & Direct-\tabfm & 0.067 & 0.041 & 0.025 & 0.024 & 0.011 & \E{8.3}{-3} & \E{8.3}{-3}$^\dagger$ \\
  & Latent-\tabfm & 0.081 & 0.057 & 0.035 & 0.032 & 0.019 & 0.017 & 0.016$^\dagger$ \\
  & LSTM & \E{2.0}{-3} & \E{5.9}{-4} & \EB{8.0}{-5} & \EB{8.8}{-5} & \EB{1.1}{-4} & \EB{9.6}{-5} & \EB{2.5}{-3}$^\dagger$ \\
  & MLP & \EB{2.8}{-4} & \EB{2.9}{-4} & \E{2.7}{-4} & \E{3.3}{-4} & \E{5.3}{-4} & \E{1.5}{-3} & \E{4.0}{-3}$^\dagger$ \\
  & LSTM-few-shot & 0.23 & 0.13 & 0.081 & 0.047 & 0.027 & 0.024 & 0.016$^\dagger$ \\
  & MLP-few-shot & 0.087 & 0.097 & 0.11 & 0.11 & 0.13 & 0.12 & 0.18$^\dagger$ \\
\midrule
\multirow{11}{*}{HalfCheetah}
  & Persistence & 1.12 & 1.14 & 1.12 & 1.13 & 1.11 & 1.12 & 1.16 \\
  & Ridge & 13.0 & 2.52 & 0.48 & 0.24 & 0.17 & 0.13 & 0.15 \\
  & \new{GP} & \new{1.47} & \new{1.17} & \new{0.59} & \new{0.20} & \new{0.13} & \new{0.080} & \new{0.088} \\
  & \new{Kernel ridge} & \new{1.45} & \new{1.19} & \new{0.80} & \new{0.55} & \new{0.36} & \new{0.20} & \new{0.19} \\
  & \new{$k$-NN} & \new{1.44} & \new{1.08} & \new{0.76} & \new{0.55} & \new{0.42} & \new{0.27} & \new{0.28} \\
  & Direct-\tabfm & 1.38 & 0.93 & 0.41 & 0.20 & 0.13 & 0.076 & 0.067 \\
  & Latent-\tabfm & 1.39 & 1.08 & 0.59 & 0.36 & 0.23 & 0.14 & 0.13 \\
  & LSTM & \EB{7.3}{-3} & \EB{6.6}{-3} & \EB{6.6}{-3} & \EB{4.2}{-3} & \EB{6.2}{-3} & \EB{6.5}{-3} & \EB{5.8}{-3} \\
  & MLP & 0.011 & \E{1.0}{-2} & 0.016 & 0.017 & 0.024 & 0.034 & 0.055 \\
  & LSTM-few-shot & 2.52 & 1.97 & 1.22 & 0.78 & 0.49 & 0.21 & 0.17 \\
  & MLP-few-shot & 1.31 & 1.11 & 0.93 & 1.00 & 0.98 & 1.02 & 1.09 \\
\bottomrule
\end{tabular}
\end{table}

\begin{table}[h]
\centering
\caption{\textbf{Paired bootstrap ratios} of Direct-\tabfm's mean error to each baseline's (values $<1$ favour Direct-\tabfm), with 95\% confidence intervals. $^\ddagger$: no interval (single episode). \new{Rollouts at all $H$ use the same start points.} --: not evaluated; the few-shot baselines are compared on one-step prediction only.}
\label{tab:boot}
\scriptsize
\setlength{\tabcolsep}{2.5pt}
\resizebox{\linewidth}{!}{\begin{tabular}{llccccc}
\toprule
Dataset & Setting & vs.\ Ridge & vs.\ Persistence & vs.\ LSTM-few-shot & \new{vs.\ GP} & vs.\ LSTM (full) \\
\midrule
\multirow{6}{*}{Lorenz}
 & $H{=}5$ & 3.33 [2.48, 5.47] & 0.012 [0.007, 0.017] & 0.47 [0.19, 0.90] & \new{359 [106, 5390]} & 0.29 [0.23, 0.41] \\
 & $H{=}20$ & 0.13 [0.082, 0.22] & 0.003 [0.003, 0.003] & 0.049 [0.035, 0.077] & \new{866 [601, 1480]} & 50.2 [32.5, 101] \\
 & $H{=}100$ & 0.003 [0.001, 0.006] & 0.001 [7.7e-04, 0.002] & 0.039 [0.023, 0.062] & \new{729 [517, 998]} & 1.39 [1.05, 2.29] \\
 & Rollout, $H{=}20$ & \new{0.59 [0.46, 1.10]} & \new{1.30 [1.15, 1.47]} & -- & \new{8.02 [6.41, 16.5]} & \new{15500 [10100, 22200]} \\
 & Rollout, $H{=}100$ & \new{0.096 [0.043, 0.12]} & \new{0.51 [0.12, 0.80]} & -- & \new{88.7 [77.2, 1610]} & \new{6050 [1190, 12900]} \\
 & Rollout, $H{=}500$ & \new{0.002 [3.0e-04, 0.005]} & \new{0.002 [2.6e-04, 0.005]} & -- & \new{23900 [10900, 27800]} & \new{18.2 [2.73, 47.7]} \\
\midrule
\multirow{6}{*}{LDS}
 & $H{=}5$ & 4.55 [3.55, 6.03] & 0.83 [0.68, 1.02] & 0.80 [0.65, 1.12] & \new{1.69 [1.42, 2.12]} & 18.1 [13.5, 22.6] \\
 & $H{=}20$ & 1.34 [1.17, 1.50] & 0.092 [0.071, 0.12] & 0.16 [0.12, 0.23] & \new{1.07 [0.92, 1.25]} & 1.89 [1.54, 2.31] \\
 & $H{=}100$ & 1.14 [1.01, 1.35] & 0.051 [0.039, 0.068] & 0.092 [0.071, 0.14] & \new{1.16 [1.01, 1.40]} & 1.10 [0.95, 1.41] \\
 & Rollout, $H{=}20$ & \new{2.87 [1.58, 6.83]} & \new{2.03 [1.32, 2.87]} & -- & \new{1.49 [0.88, 3.09]} & \new{2.68 [1.55, 4.71]} \\
 & Rollout, $H{=}100$ & \new{1.90 [1.26, 2.87]} & \new{1.33 [0.98, 1.61]} & -- & \new{1.61 [1.21, 2.24]} & \new{1.76 [1.26, 2.48]} \\
 & Rollout, $H{=}500$ & \new{1.35 [1.05, 1.90]} & \new{0.87 [0.83, 0.93]} & -- & \new{1.35 [1.09, 1.83]} & \new{1.15 [0.86, 1.47]} \\
\midrule
\multirow{6}{*}{Pendulum}
 & $H{=}5$ & 6.45 [3.78, 10.4] & 0.39 [0.26, 0.53] & 0.66 [0.55, 0.82] & \new{4.70 [1.79, 24.2]} & 3.24 [2.02, 5.58] \\
 & $H{=}20$ & 1.45 [0.86, 2.16] & 0.012 [0.006, 0.021] & 0.018 [0.009, 0.033] & \new{3410 [1750, 16300]} & 293 [155, 490] \\
 & $H{=}100$ & 21.3 [7.91, 37.3] & 0.002 [8.3e-04, 0.004] & 0.008 [0.003, 0.014] & \new{953 [533, 1780]} & 8.30 [2.85, 17.3] \\
 & Rollout, $H{=}20$ & \new{4.64 [4.27, 5.58]} & \new{1.35 [0.95, 1.56]} & -- & \new{28900 [19900, 131000]} & \new{26300 [14400, 48800]} \\
 & Rollout, $H{=}100$ & \new{6.58 [3.08, 13.9]} & \new{0.002 [8.7e-04, 0.004]} & -- & \new{975 [576, 1730]} & \new{35.5 [25.6, 45.4]} \\
 & Rollout, $H{=}500$ & \new{36.3 [1.75, 172]} & \new{6.1e-04 [5.3e-05, 0.001]} & -- & \new{1430 [205, 1870]} & \new{11.9 [0.68, 35.4]} \\
\midrule
\multirow{6}{*}{Hopper}
 & $H{=}5$ & 0.11 [0.062, 0.22] & 0.79 [0.72, 0.86] & 0.29 [0.23, 0.37] & \new{0.70 [0.65, 0.75]} & 32.8 [27.9, 38.2] \\
 & $H{=}20$ & 0.48 [0.24, 0.93] & 0.27 [0.20, 0.34] & 0.32 [0.24, 0.44] & \new{0.67 [0.55, 0.83]} & 319 [187, 588] \\
 & $H{=}100$ & 0.55 [0.44, 0.64] & 0.12 [0.076, 0.17] & 0.42 [0.35, 0.49] & \new{0.56 [0.47, 0.64]} & 102 [79.2, 132] \\
 & Rollout, $H{=}20$ & \new{0.27$^\ddagger$} & \new{3.49$^\ddagger$} & -- & \new{1.22$^\ddagger$} & \new{2960$^\ddagger$} \\
 & Rollout, $H{=}100$ & \new{0.059$^\ddagger$} & \new{0.46$^\ddagger$} & -- & \new{0.35$^\ddagger$} & \new{392$^\ddagger$} \\
 & Rollout, $H{=}500$ & \new{0.56$^\ddagger$} & \new{0.26$^\ddagger$} & -- & \new{0.15$^\ddagger$} & \new{218$^\ddagger$} \\
\midrule
\multirow{6}{*}{HalfCheetah}
 & $H{=}5$ & 0.11 [0.057, 0.32] & 1.24 [1.20, 1.27] & 0.55 [0.53, 0.56] & \new{0.94 [0.92, 0.96]} & 190 [138, 289] \\
 & $H{=}20$ & 0.85 [0.79, 0.91] & 0.37 [0.35, 0.38] & 0.34 [0.32, 0.36] & \new{0.70 [0.66, 0.73]} & 62.4 [46.1, 88.6] \\
 & $H{=}100$ & 0.76 [0.71, 0.81] & 0.12 [0.10, 0.13] & 0.26 [0.23, 0.29] & \new{0.96 [0.90, 1.04]} & 20.6 [17.3, 25.5] \\
 & Rollout, $H{=}20$ & \new{0.026 [0.008, 0.56]} & \new{2.34 [1.69, 3.04]} & -- & \new{0.76 [0.51, 1.17]} & \new{6.00 [4.09, 8.33]} \\
 & Rollout, $H{=}100$ & \new{0.73 [0.64, 0.80]} & \new{0.15 [0.11, 0.20]} & -- & \new{0.96 [0.80, 1.19]} & \new{0.39 [0.25, 0.59]} \\
 & Rollout, $H{=}500$ & \new{0.73 [0.62, 0.84]} & \new{0.13 [0.074, 0.19]} & -- & \new{0.76 [0.51, 1.00]} & \new{0.33 [0.17, 0.55]} \\
\bottomrule
\end{tabular}}
\end{table}

\begin{table}[h]
\centering
\caption{\textbf{20-step rollout state MSE vs.\ context length} \new{for all methods, from the same start points (final step). Bold: within 5\% of the lowest per dataset. Hopper uses 36 starts from a single episode.}}
\label{tab:rollfull}
\scriptsize
\setlength{\tabcolsep}{4pt}
\begin{tabular}{lccccc}
\toprule
\new{Method} & \new{Lorenz} & \new{LDS} & \new{Pendulum} & \new{Hopper} & \new{HalfCheetah} \\
\midrule
\multicolumn{6}{l}{\new{\textit{Independent of $H$}}} \\
\new{Persistence} & \new{\E{9.3}{1}} & \new{\E{2.5}{-2}} & \new{\E{4.6}{-1}} & \new{\E{7.6}{0}} & \new{\E{1.9}{0}} \\
\new{LSTM (full data)} & \new{\E{7.8}{-3}} & \new{\E{1.9}{-2}} & \new{\E{2.4}{-5}} & \new{\EB{9.0}{-3}} & \new{\E{7.5}{-1}} \\
\new{MLP (full data)} & \new{\E{2.2}{-1}} & \new{\E{3.0}{-2}} & \new{\E{1.8}{-4}} & \new{\E{1.1}{-1}} & \new{\E{6.3}{-1}} \\
\midrule
\multicolumn{6}{l}{\new{\textit{$H{=}20$}}} \\
\new{Ridge} & \new{\E{2.0}{2}} & \new{\E{1.8}{-2}} & \new{\E{1.3}{-1}} & \new{\E{9.7}{1}} & \new{\E{1.7}{2}} \\
\new{GP} & \new{\E{1.5}{1}} & \new{\E{3.4}{-2}} & \new{\E{2.2}{-5}} & \new{\E{2.2}{1}} & \new{\E{5.9}{0}} \\
\new{Kernel ridge} & \new{\E{2.0}{2}} & \new{\E{2.3}{-2}} & \new{\E{1.8}{0}} & \new{\E{2.1}{1}} & \new{\E{2.4}{0}} \\
\new{$k$-NN} & \new{\E{1.9}{2}} & \new{\E{4.6}{-2}} & \new{\E{2.1}{0}} & \new{\E{3.2}{1}} & \new{\E{9.4}{0}} \\
\new{\tabfm} & \new{\E{1.2}{2}} & \new{\E{5.1}{-2}} & \new{\E{6.2}{-1}} & \new{\E{2.6}{1}} & \new{\E{4.5}{0}} \\
\midrule
\multicolumn{6}{l}{\new{\textit{$H{=}100$}}} \\
\new{Ridge} & \new{\E{4.9}{2}} & \new{\E{1.8}{-2}} & \new{\E{1.3}{-4}} & \new{\E{6.0}{1}} & \new{\E{4.1}{-1}} \\
\new{GP} & \new{\E{5.3}{-1}} & \new{\E{2.1}{-2}} & \new{\E{8.6}{-7}} & \new{\E{1.0}{1}} & \new{\E{3.1}{-1}} \\
\new{Kernel ridge} & \new{\E{5.9}{1}} & \new{\E{3.7}{-2}} & \new{\E{5.8}{-2}} & \new{\E{7.8}{0}} & \new{\E{3.9}{-1}} \\
\new{$k$-NN} & \new{\E{1.3}{2}} & \new{\E{3.8}{-2}} & \new{\E{6.4}{-2}} & \new{\E{2.8}{1}} & \new{\E{1.3}{0}} \\
\new{\tabfm} & \new{\E{4.7}{1}} & \new{\E{3.3}{-2}} & \new{\E{8.4}{-4}} & \new{\E{3.5}{0}} & \new{\E{2.9}{-1}} \\
\midrule
\multicolumn{6}{l}{\new{\textit{$H{=}500$}}} \\
\new{Ridge} & \new{\E{8.4}{1}} & \new{\EB{1.6}{-2}} & \new{\E{7.7}{-6}} & \new{\E{3.5}{0}} & \new{\E{3.4}{-1}} \\
\new{GP} & \new{\EB{5.9}{-6}} & \new{\EB{1.6}{-2}} & \new{\EB{2.0}{-7}} & \new{\E{1.3}{1}} & \new{\E{3.2}{-1}} \\
\new{Kernel ridge} & \new{\E{6.2}{-1}} & \new{\E{4.0}{-2}} & \new{\E{3.7}{-2}} & \new{\E{5.6}{0}} & \new{\E{2.7}{-1}} \\
\new{$k$-NN} & \new{\E{1.4}{0}} & \new{\E{2.7}{-2}} & \new{\E{2.7}{-2}} & \new{\E{2.2}{1}} & \new{\E{5.7}{-1}} \\
\new{\tabfm} & \new{\E{1.4}{-1}} & \new{\E{2.2}{-2}} & \new{\E{2.8}{-4}} & \new{\E{2.0}{0}} & \new{\EB{2.5}{-1}} \\
\bottomrule
\end{tabular}
\end{table}

\begin{table}[h]
\centering
\caption{\textbf{Inference time per one-step prediction on Hopper} (mean wall-clock seconds\new{; GP, kernel ridge and $k$-NN on CPU, all others on a single GPU}).}
\label{tab:latency}
\scriptsize
\begin{tabular}{lcc}
\toprule
Method & $H{=}20$ & $H{=}500$ \\
\midrule
Persistence & \E{5.9}{-6} & \E{5.5}{-6} \\
Ridge & \E{8.0}{-4} & \E{9.0}{-4} \\
\new{$k$-NN} & \new{\E{5.7}{-3}} & \new{\E{7.0}{-3}} \\
\new{Kernel ridge} & \new{\E{6.5}{-3}} & \new{0.052} \\
\new{GP} & \new{0.45} & \new{15.8} \\
LSTM (full data) & \E{6.3}{-3} & \E{2.2}{-3} \\
MLP (full data) & \E{1.9}{-4} & \E{2.4}{-4} \\
MLP-few-shot & 0.034 & 0.080 \\
LSTM-few-shot & 0.22 & 2.39 \\
Direct-\tabfm & 3.18 & 3.58 \\
Latent-\tabfm & 3.94 & 5.21 \\
\bottomrule
\end{tabular}
\end{table}

\FloatBarrier
\section{Extended Related Work}
\label{app:related}

\paragraph{Tabular foundation models.}
Tabular foundation models are pretrained on large collections of synthetic or real tabular datasets and perform in-context regression or classification at test time without task-specific fine-tuning. \tabpfn~\citep{tabpfnv2} follows the prior-data fitted networks (PFN) framework~\citep{pfn} and is trained on datasets sampled from SCMs. At inference, it conditions on a context of input-output pairs treated as a dataset and predicts queries in a single forward pass. Related models include TabICL~\citep{tabicl}, which scales tabular in-context learning to large datasets, and language-model-based approaches such as Tabula-8B~\citep{tabula8b}. TabPFN-TS~\citep{tabpfnts} extends this paradigm to time-series forecasting, linking tabular in-context learning with dynamical system prediction via past-to-future regression. Recent work has also used \tabpfn~and related models as gradient-free Q-function approximators for reinforcement learning~\citep{icrrl,tabql}, and extended TabPFN-based forecasting to multivariate series by presenting all channels to a single model~\citep{jayawardhana2026zeroshot}. In contrast, our work targets multivariate, action-conditioned state prediction using transition pairs. Unlike sequence models trained autoregressively, tabular foundation models operate over sets of examples and perform inference entirely in-context without gradient updates.

\paragraph{World models and learned dynamics.}
Learning predictive models of environment dynamics for planning has a long history~\citep{worldmodels,moerland2023}. Classical model-based RL methods such as PETS~\citep{chua2018deep} and MBPO~\citep{janner2019trust} learn explicit dynamics models using ensembles or neural networks trained on environment-specific data. Recurrent State Space Models (RSSMs)~\citep{rssm} form the basis of Dreamer-style agents, combining learned priors, posterior corrections, and decoders for pixel reconstruction. DreamerV3~\citep{dreamerv3} scales this approach with improved latent representations and normalization, while TD-MPC2~\citep{tdmpc2} uses latent dynamics for model-predictive control. More broadly, methods such as JEPA~\citep{jepa} and Cosmos~\citep{cosmos} learn latent or video-based world models from large-scale data, and world action models~\citep{wang2026world} aim to unify prediction and action for embodied agents. These approaches rely on gradient-based optimization and either environment-specific or large-scale interaction data. A known difficulty for all of them is that the one-step training objective does not directly optimize multi-step or control performance~\citep{talvitie2014model,venkatraman2015improving,lambert2020objective}.

\paragraph{In-context learning and function conditioning.}
A growing line of work studies in-context and fast-adaptation methods for dynamical systems. \citet{benechehab2025zeroshot} show that pretrained LLMs can predict the dynamics of continuous control tasks in-context, and use the resulting model for policy evaluation and data-augmented off-policy RL. OSWM~\citep{ferreira2024oneshot} follows the PFN framework but uses a prior of random neural networks designed to generate dynamics, and trains policies inside the resulting in-context world model. Transformers meta-trained on synthetic system classes have also been proposed for in-context system identification~\citep{forgione2023system} and as foundation models for control-relevant dynamical systems~\citep{ziegler2024foundation}, and LLMs exhibit an in-context scaling law on dynamical systems~\citep{liu2024llms}. Neural Context Flows~\citep{ncf} and related meta-learning approaches enable rapid adaptation at test time via gradient updates on environment-specific context vectors. In-context learning has been analyzed on simple function classes such as linear regression~\citep{icl_dynamics}, and time-series foundation models have been applied to zero-shot forecasting of chaotic systems~\citep{zhang2025zeroshot}. Conditional Neural Processes~\citep{cnp} and Attentive Neural Processes~\citep{anp} condition on context sets to model stochastic functions but require training on task distributions. In contrast, we investigate whether a general-purpose tabular foundation model pretrained on SCM-generated data can perform in-context dynamics prediction without task-specific training.

\end{document}